\documentclass[sn-mathphys]{sn-jnl}

\jyear{2021}%

\theoremstyle{thmstyleone}%
\theoremstyle{thmstyletwo}%

\theoremstyle{thmstylethree}%

\begin{document}

\title[Research on Parallel SVM Algorithm Based on Cascade SVM]{Research on Parallel SVM Algorithm Based on Cascade SVM}


\author[1]{\fnm{YiCheng} \sur{Liu}}\email{yc\_rhonin@qq.com}
\author*[1]{\fnm{XiaoYan} \sur{Liu}}\email{906384457@qq.com}



\affil[1]{\orgdiv{Faculty of Information Engineering and Automation}, \orgname{Kunming University of Science and Technology}, \orgaddress{\city{Kunming}, \postcode{650504}, \country{China}}}




\abstract{
    Cascade SVM (CSVM) can group datasets and train subsets in parallel, which greatly reduces the training time and memory consumption. However, the model accuracy obtained by using this method has some errors compared with direct training. In order to reduce the error, we analyze the causes of error in grouping training, and summarize the grouping without error under ideal conditions. A Balanced Cascade SVM (BCSVM) algorithm is proposed, which balances the sample proportion in the subset after grouping to ensure that the sample proportion in the subset is the same as the original dataset. At the same time, it proves that the accuracy of the model obtained by BCSVM algorithm is higher than that of CSVM. Finally, two common datasets are used for experimental verification, and the results show that the accuracy error obtained by using BCSVM algorithm is reduced from 1\% of CSVM to 0.1\%, which is reduced by an order of magnitude.
}




\maketitle

\section{Introduction}\label{sec1}

Support Vector Machines (SVM) \cite{vapnik1995} is a very powerful tool in the field of classification and regression, but its calculation and storage requirements grow exponentially with the increase of training samples. The core of support vector machine is a Quadratic Programming (QP) problem. A key factor limiting the application of SVM in large sample problems is that the super-large Quadratic Programming problems induced by SVM training cannot be solved by standard Quadratic Programming methods. SMO algorithm (Sequential Minimal Optimization (SMO)\cite{SMO} is a large-scale SVM training algorithm. Its advantage is that each step of Optimization only deals with two samples, and it can be solved quickly by analytical methods. Therefore, it is faster than other algorithms in computing large sample sets. However, when the number of samples increases further, the increase of its computational complexity is still much faster than the increase of the number of samples, so it is difficult to meet the actual demand.

Since support vector machines only rely on support vectors in the training sample set to make decisions, non-support vectors have no effect and are just useless information. Removing them does not affect the training results at all. It is the processing of such useless information that occupies the main time of the algorithm, so one of the ideas to speed up SVM training is "chunking". The so-called block is to divide the original sample set into many subsets, and only one subset is optimized by the SMO algorithm in each step. The feature that non-support vectors can be removed after training is used to accelerate the training speed.

Wu Xiang et al. \cite{wuxiang} improved the idea of block algorithm and proposed the block SMO algorithm. The improved algorithm has a very obvious advantage in speed for the large sample problem with a small support vector set. Graf et al. \cite{cascadeSVM} proposed the cascade support vector machine, which is also an algorithm based on the idea of partitioning. In this algorithm, the data sets are randomly grouped, the subsets after grouping are trained in parallel, and the support vectors obtained by training are combined in pairs. After the combination, the training is repeated to obtain a final training result. Finally, whether the final result is regrouped is determined according to the conditions. In recent years, the cascade support vector machine algorithm has been combined with distributed computing for training, making full use of the advantages of distributed computing that can flexibly schedule multiple computing nodes, greatly reducing the training time. For example, Zhang Pengxiang\cite{mapreduce}, Liu Zeshan \cite{spark}, Nie\cite{hadoop}, Bai Yusxin \cite{flink} and others have implemented corresponding parallel training algorithms based on cascade support vector machines on MapReduce, Spark, Hadoop and Flink platforms respectively. However, because the data set is randomly segmented based on the cascaded support vector machine, there is a certain probability that the real global support vector of the whole sample set will be removed during the training of the subset, and the final model will inevitably have certain errors compared with the direct training.

This paper analyzes the situation of global support vector reservation, proposes Balanced Cascade Support Vector Machine, tries to use BCSVM algorithm to reduce the error between CSVM and direct training, and proves mathematically that BCSVM algorithm can preserve global support vector with higher probability. The organization of this paper is as follows: Chapter \ref{sec2} briefly introduces support vector machines and cascade support vector machines; Chapter \ref{sec3} presents BCSVM and explains it in detail in pseudocode. Chapter \ref{sec4} proves the validity of BCSVM in a probabilistic way. Chapter \ref{sec5} further verifies the utility of BCSVM through experiments. The last chapter summarizes the article.

\section{Preliminaries}\label{sec2}
\subsection{Support Vector Machines}
Support vector machine is a supervised learning model that can be used for classification and is one of the most famous machine learning algorithms\cite{engineering2019}. The principle is to find the classification hyperplane with the maximum edge distance in the feature space to determine the classification of new data.

Given a binary data set with $n$ data points $\mathcal{I}$, which contains $n$ data points $x_i$ and the corresponding label $y_i \in \{+1,-1\}$, we call the label $y_i=+1$ samples positive samples, and the others negative samples. All of the samples $x_i$ in the data set are a $d$ dimensional vector, which is $x_i \in \mathbb{R}^d$. Support vector machines are all about finding a $d-1$ dimensional hyperplane to distinguish between two categories in a sample. However, in Euclidean space, real-world data tends to be linearly indivisible. Therefore, in practice, in order to make the two categories distinguishable by hyperplane, data is often mapped to a higher dimensional space $\phi: \mathbb{R}^d \rightarrow \mathbb{R}^p (d \leq p)$. This method is also called kernel trick, which was first proposed by Aizerman et al.\cite{aizerman1964}, and then applied to support vector machines by Boser et al.\cite{boser1992}

\subsection{Cascade SVM}
\begin{figure}[ht]
    \centering
    \includegraphics[scale=0.4]{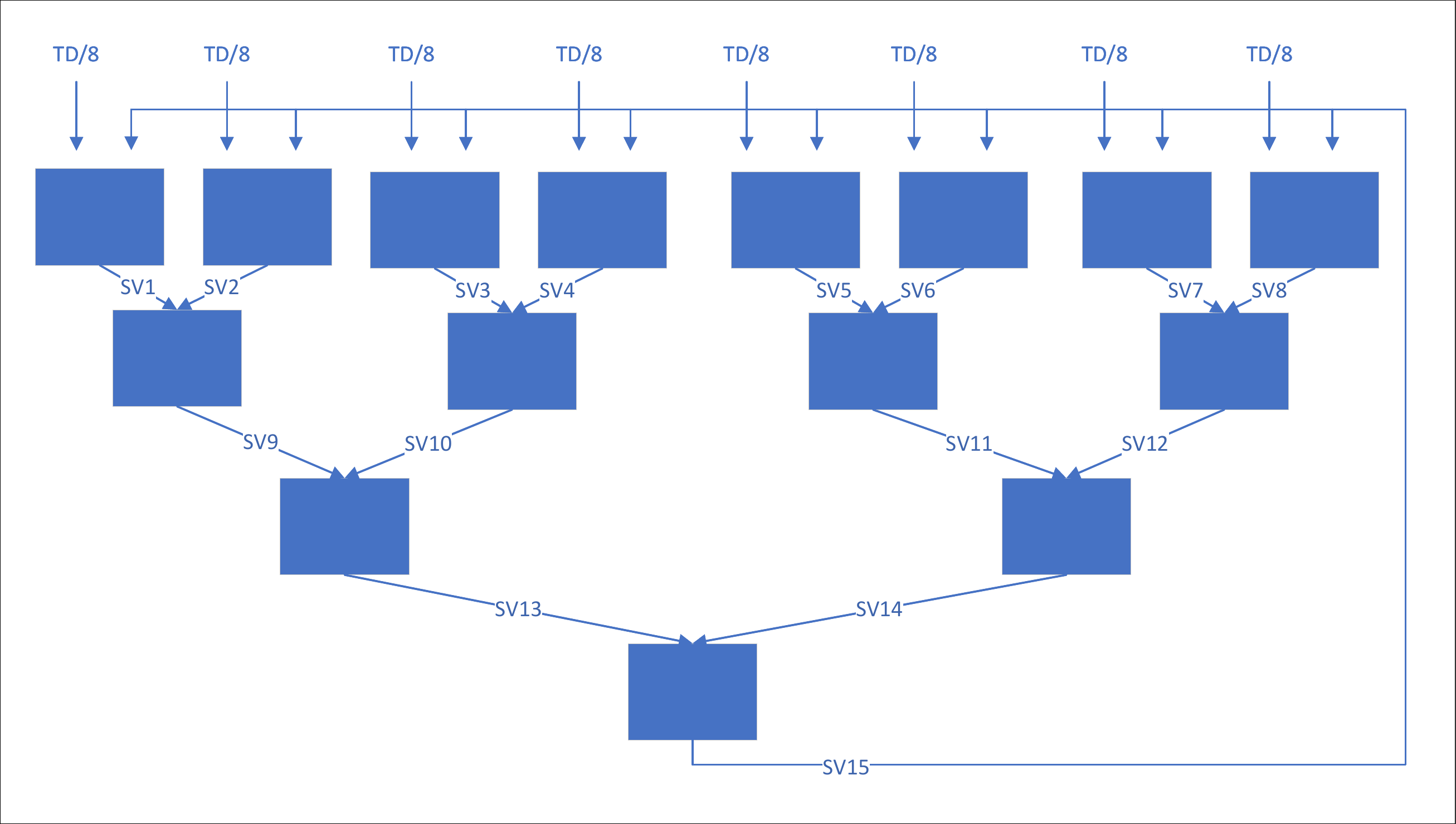}
    \caption{Schematic diagram of cascaded support vector machine architecture}
    \label{figure:1}

\end{figure}

The architecture of cascaded support vector machines is shown in Figure \ref{figure:1}. Firstly, the algorithm divides the whole dataset into several subsets, trains each subset in parallel in the first layer, and combines the obtained support vectors in pairs as the input of the next layer. In this architecture, one SVM does not need to process the entire training set. If the first layers can filter out most of the non-support vectors, the last layer needs to process very few samples.

The advantage of this algorithm is that each subset only needs to process part of the data, and the training among each subset is independent, which can perform parallel computing well. Therefore, compared with the direct training of the whole data set, cascaded support vector machine can greatly accelerate the training speed. However, since the grouping method is random grouping, there is a high probability of losing the global support vector. That is, there is still a certain gap between the model finally obtained and the model directly trained. And because it adopts binary cascade, there are multiple merge behaviors. If most of the merged support vectors are global support vectors, only a small number of samples can be removed after the merged training, and the training efficiency will be very low. To solve this problem, The SP-SVM proposed by Liu Zeshan et al\cite{spark} modified the cascade mode in the distributed cluster environment. They integrated the support vectors obtained from each layer of training into a training set, and then conducted group training on the integrated training set. This cascading mode effectively solves the problem of low utilization of cluster resources in a distributed cluster environment. The grouped parallel SVM algorithm is especially suitable for data sets with large number of samples and small support vectors. In this case, the training time and memory consumption show obvious advantages, because most samples can be filtered each time. However, its disadvantages are also obvious. In the case that the support vector occupies the majority of the data set, each segmentation basically cannot filter the sample, and in extreme cases, its training speed will even be slower than direct training.

\section{Balance Cascade SVM}\label{sec3}

Balanced Cascaded Support Vector Machine adopts the idea of group cascade in cascaded support vector machine, combined with SP-SVM in the cascade of each layer of training results, and modified the group mode of cascaded support vector machine on this basis.

The difference between the balanced cascaded support vector machine and the cascaded support vector machine lies in the grouping method. The grouping method adopted by the cascaded support vector machine is random grouping, while the balanced cascaded support vector machine balances the sample proportion in the subset after grouping, that is, the sample proportion in each subset is the same as the original data set. Its pseudocode is shown in algorithm \ref{algorithm:1}. Where classifyNum is the group number of each layer, for example, the number group [4,1] indicates that the first layer is divided into four subsets, and the second layer only has one subset. It should be noted that the last element in classifyNum must is 1, so as to obtain a unique training model. The algorithm firstly separates the training set according to the type of samples, and then divides all kinds of samples into each subset in a balanced way. After that, we train the subsets in parallel and integrate the support vectors, and then train the support vectors in groups, and finally get a unique model.

Compared with CSVM, BCSVM only modified the grouping method of subsets, and the time complexity of the original SMO algorithm was between $O(n^2)$ and $O(n^3)$, while the time complexity after grouping was $O(k \times (\frac{n}{k})^{2\sim 3}) = O(n^{2\sim 3})$ ($n$ was the number of samples, $k$ was the number of groups), so there was no essential change in its complexity. However, in the application, each subset only needs to process the data in the subset, so the actual training time can be greatly reduced. Generally, in order to speed up the training speed, the program will store the kernel function matrix in memory, so its space complexity is $O(n^2)$. In the same way as the time complexity, the space complexity does not change substantially after grouping, only changing $O(n^2)$ to $O(\frac{n^2}{k})$. However, most SVM training tools adopt certain strategies to limit memory occupation, for example, LibSVM\cite{LIBSVM} adopts LRU(Least Recently Used) algorithm to control memory occupation.

\begin{algorithm}[ht]
    \caption{Pseudo code of BCSVM}
    \label{algorithm:1}
    \begin{algorithmic}[1]
        \Require trainDataSet,classifyNum
        \Ensure Model

        \For {$num$ in classifyNum}
        \State [sameClassDataSet] = clasifyByClass(trainDataSet)
        \State subDataSets = []
        \For {$i=0$ to $num$}
        \State [sameClassDataNum] = \emph{count the number of every class samples}
        \State subDataSet = \emph{take $\frac{sameClassDataNum}{num}$ samples of each category as a subset}
        
        \State subDataSets.push(subDataSet)
        \EndFor

        \State SV = []
        \For {subDataSet in subDataSets}
            \State SVi = SVM(subDataSet).getSVs()
            \State SV.extend(SVi)
        \EndFor 
        \State trainDataSet = SV
        \EndFor \\
        \Return SVM(trainDataSet)

    \end{algorithmic}

\end{algorithm}

\section{Prove}\label{sec4}
\begin{figure}
    \centering
    \includegraphics[scale=1]{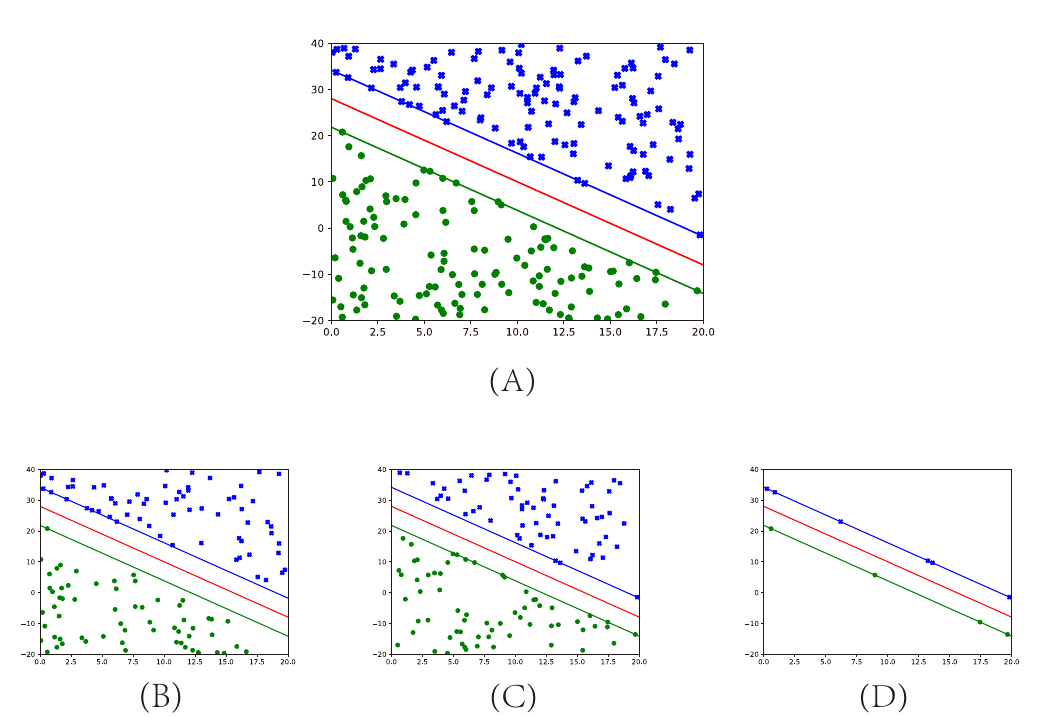}
    \caption{Group training for linear fractionable data sets. The middle of the three lines is the separation hyperplane, and the remaining lines are the constraint boundary of the sample}
    \label{figure:2}
\end{figure}

Through the research and analysis of SVM training process, there are the following situations in which global support vector is not removed in grouping training:

1. There is no noise in the dataset. Noise is linearly indivisible data. If the data does not exist, then the dataset is linearly separable. Suppose $\Omega$ represents the entire data set, $S$ represents the support vector, $C=\Omega - S$ represents any sample other than the support vector, and $d(x)$ represents the geometric distance between the sample and the detached hyperplane. There are clearly
\begin{gather*}
    \forall s \in S,\quad \forall c \in C \\ d(s) < d(c)
\end{gather*}
In other words, if the global support vector exists, the global support vector must be the sample closest to the separated hyperplane, that is, it must not be discarded. At the same time, it is shown that if the whole dataset is linearly separable, the optimal training result can be obtained as long as any subset of grouping contains multiple types of samples. As shown in Figure \ref{figure:2}, the data set in A is linearly separable, BC shows the training result of randomly dividing the data set in A into two subsets, and D is the retraining of the support vector obtained by summarizing BC training. It can be seen that after BC filtering, d only needs to train fewer samples, and the training results obtained in D are completely consistent with direct training of A. Meanwhile, it was also noted that the training results of BC were consistent with those of A. This happens only when BC contains global support vectors. At the beginning, it is not clear that those samples belong to global support vectors, so the support vectors obtained by BC training should be trained again to get a final result D.


\begin{figure}[ht]
    \centering
    \includegraphics[scale=0.7]{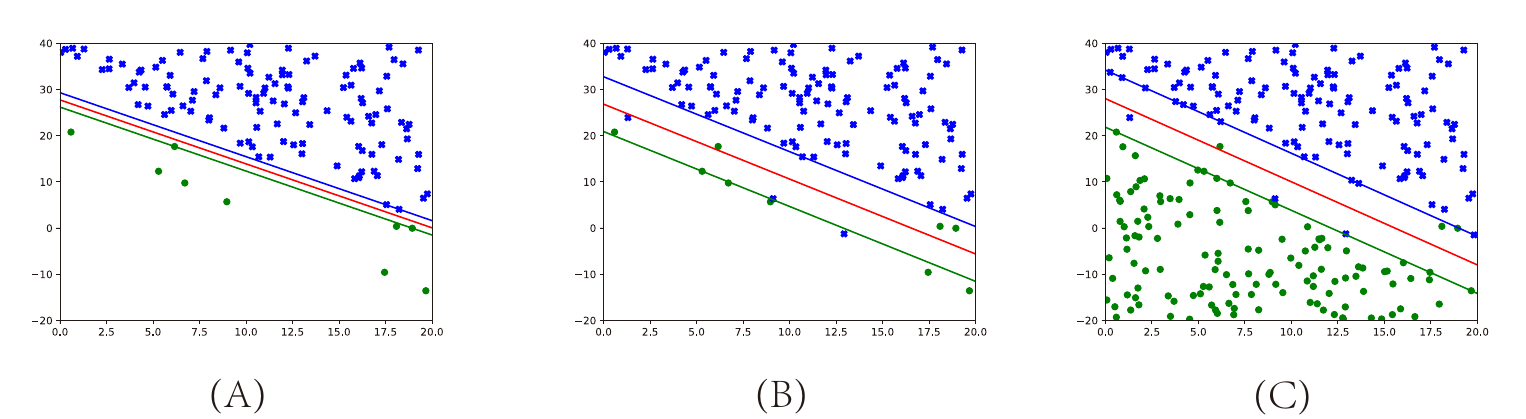}
    \caption{The case that the linear non-fractional data set retains the global support vector under grouping training}
    \label{figure:3}

\end{figure}

2.Support vector and noise exist simultaneously. Taking the positive sample as an example, when the global positive sample support vector, positive noise (misclassified positive sample) and negative noise (misclassified negative sample) exist at the same time in the subset, the global positive sample support vector will have great probability retention. If the negative noise does not exist, the global positive sample support vector is likely to be lost. See Figure \ref{figure:3}. Figure C represents the training results of the linear non-fractional data set, and the samples between the uppermost and bottommost lines are noise points. The dataset in AB is a subset of the dataset in C. In Figure A, the global positive sample support vector and positive noise exist simultaneously, but there is no negative sample. After training, the global positive sample support vector is lost. Figure B also contains global positive sample support vector, positive noise point and negative noise point, and most of the global positive sample support vector is retained after training.

Only in these two cases, the global support vector has a great probability of retention, and the other grouping methods have a great probability of losing the global support vector. In this paper, the samples in the data set are divided into support vector, noise point and ordinary sample. Before the training, only all positive samples and all negative samples are known, and other information about the samples is not known. If grouping is regarded as sampling of samples, both CSVM and BCSVM have the same sampling types of global support vector reservation. The difference lies in that BCSVM has fewer sampling methods than CSVM, so the probability of global support vector reservation of BCSVM is higher than that of CSVM. Taking the positive sample global support vector retention probability in dichotomies as an example, the proof process is as follows, and the specific significance of symbols is shown in Table \ref{table:1}.

\begin{table}[h]
    \begin{center}
        \caption{ Symbols and meanings}
        \label{table:1}
        \begin{tabular}{@{}ll@{}}
            \toprule
            symbol & mean                            \\
            \hline
            pSv    & positive support vector         \\
            \hline
            nSv    & negative support vector         \\
            \hline
            pN     & positive noisy                  \\
            \hline
            nN     & negative noisy                  \\
            \hline
            pDS    & common positive data set        \\
            \hline
            nDS    & common negative data set        \\
            \hline
            pT     & all positive samples=pSv+pDS+pN \\
            \hline
            nT     & all negative samples=nSv+nDS+nN \\
            \hline
            T      & all samples=pT+nT               \\
            \hline
            m      & num of chunking                 \\
            \botrule
        \end{tabular}
    \end{center}

\end{table}

The possible sampling outcomes for case 1 are
\begin{gather*}
    C_{pSv}^{t_1} \times C_{pDS}^{t_2} \times C_{nDS+nSv}^{t_3}\\
    1 \leq t_1 \leq pSv \\
    0 \leq t_2 \leq pDS \\
    0 \leq t_3 \leq nDS \\
    t = t_1 +t_2 + t_3
\end{gather*}
Where $C_{pSv}^{t_1}$ represents $t_1$ samples randomly selected from the positive sample support vector, and $t=\frac{T}{m}$ represents the number of samples in each subset.

The possible sampling outcomes for case 2 are
\begin{gather*}
    C_{pSv}^{t_4} \times C_{pN}^{t_5}  \times C_{nN}^{t_6} \times C_{pDS}^{t_7} \times C_{nDS+nSv}^{t_8} \\
    1 \leq t_4 \leq pSv \\
    1 \leq t_5 \leq pN \\
    1 \leq t_6 \leq nN \\
    0 \leq t_7 \leq pDS \\
    0 \leq t_8 \leq nDS \\
    t = t_4+t_5+t_6+t_7+t_8
\end{gather*}
In other words, positive sample support vector, positive noise and negative noise exist simultaneously in the subset. In the case of random grouping, the sampling result of the subset is $C_T^{t}$. Therefore, in the case of random grouping, the probability of retaining the global support vector of positive samples is:
\begin{gather*}
    \frac
    {C_{pSv}^{t_1} \times C_{pDS}^{t_2} \times C_{nDS+nSv}^{t_3} + C_{pSv}^{t_4} \times C_{pN}^{t_5}  \times C_{nN}^{t_6} \times C_{pDS}^{t_7} \times C_{nDS+nSv}^{t_8} }
    {C_T^{t}}
\end{gather*}
Each subset in the BCSVM contains the same proportion of samples. So the number of positive samples $p=\frac{pT}{m}$ in each subset, the number of negative samples $n=\frac{nT}{m}$, and obviously $p+n=t$. The possible sampling result in this case is $C_{pT}^p \times C_{nT}^n$, that is, sample $p$ from all positive samples and negative sample $n$ from all negative samples. Therefore, in the case of uniform grouping, the probability of a positive sample global support vector not being removed is
\begin{gather*}
    \frac
    {C_{pSv}^{t_1} \times C_{pDS}^{t_2} \times C_{nDS+nSv}^{t_3} + C_{pSv}^{t_4} \times C_{pN}^{t_5}  \times C_{nN}^{t_6} \times C_{pDS}^{t_7} \times C_{nDS+nSv}^{t_8} }
    {C_{pT}^p \times C_{nT}^n}
\end{gather*}
because of $pT+nT=T,n+p=t$, so obviously that

$$
    C_T^t > C_{pT}^p \times C_{nT}^n
$$

\begin{figure}
    \centering
    \includegraphics[scale=0.7]{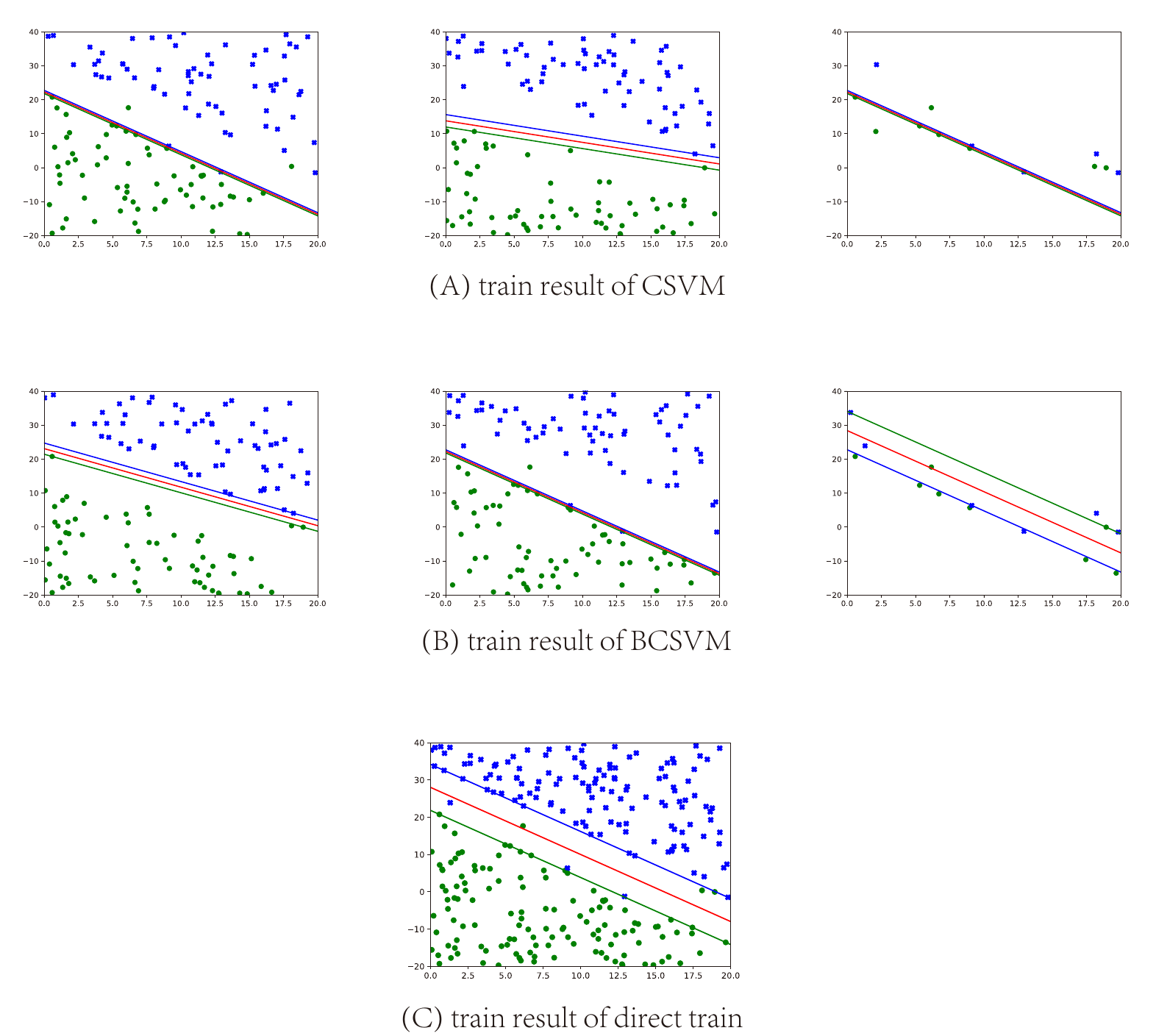}
    \caption{The influence of different algorithms on training results}
    \label{figure:4}
\end{figure}

That is, BCSVM has a higher probability of retaining the global support vector than CSVM. A brief example is shown in Figure \ref{figure:4}. ABC adopts the same data set, AB adopts CSVM and BCSVM for training respectively, and the third picture of AB shows the final training results. It can be seen that the training results obtained by using BCSVM are closer to those obtained by direct training without grouping.

\section{Experiment}\label{sec5}
The training sets used in this experiment are two standard datasets, a9a\footnote{The a9a dataset contains 213 feature values,32562 training samples and 16281 test samples. https://www.csie.ntu.edu.tw/~cjlin/libsvmtools/datasets/binary.html\#a9a} and ijcnn1\footnote{The ijcnn1 dataset contains 22 feature values,49990 training samples and 91701 test samples. https://www.csie.ntu.edu.tw/~cjlin/libsvmtools/datasets/binary.html\#ijcnn1}, both of which are dichotomous samples. The training tool for SVM uses LibSVM. Hardware configuration: 16GB RAM, 8-core 1.8GHz base speed processor.

After the data set is read into the memory, a 'shuffle' operation is performed to ensure that the original sample order of the data set will not interfere with the experiment. During the experiment, both CSVM and BCSVM divided the first data set into 8 parts, and then integrated the obtained support vector into a training set by combining the data cascade mode in SP-SVM. Finally, the set was trained to obtain the final result. The operation of CSVM is to select A sample from the data set after shuffling from the front to the back as A subset. The operation of BCSVM is to first divide the data set into a positive sample data set and negative sample data set, and select B and C samples respectively from the positive and negative sample data set to form A subset. Both data sets are trained by gaussian kernel function, and the same parameter combination is used in different grouping methods.

\begin{figure}[ht]
    
    \centering
    \includegraphics[scale=0.5]{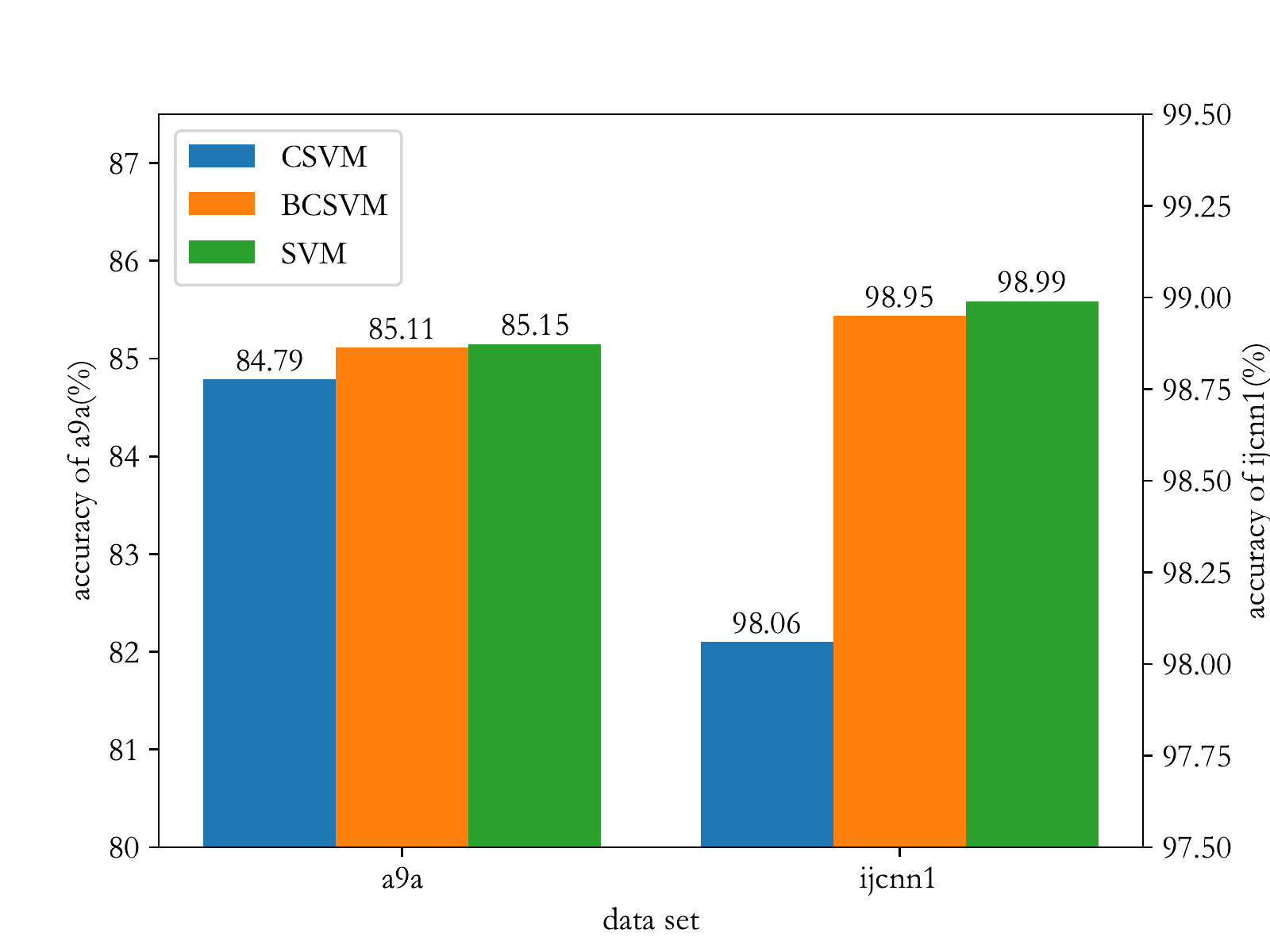}
    \caption{ The influence of different algorithms on accuracy}
    \label{figure:5}
\end{figure}

Figure \ref{figure:5} evaluates the prediction accuracy of direct training, CSVM and BCSVM in different data sets, from which it can be seen that the three different algorithms affect the prediction accuracy. The results obtained by using the same set of parameters are obviously the highest in direct training, and the accuracy of BCSVM is higher than that of CSVM. Direct training without grouping can retain all support vectors in this parameter combination, so the accuracy of both BCSVM and CSVM is obviously smaller than that. BCSVM has a higher probability of retaining a global support vector than CSVM, so the prediction accuracy of model obtained by BCSVM is higher than that of CSVM.

\section{Conclusion}\label{sec7}

In this paper, BCSVM is proposed based on CSVM and SP-SVM data cascade mode. Compared with CSVM, BCSVM retains all its characteristics, and at the same time, BCSVM can retain the global support vector with a greater probability during training. Therefore, the prediction accuracy rate of model obtained by BCSVM is higher than that of CSVM. Due to the relationship between test set and training set, it may also occur that the model accuracy obtained by BCSVM is higher than that of CSVM when the test set is used for accuracy prediction.

Due to the randomness of molecular data set partitioning, results obtained by BCSVM are more likely to be better than those obtained by CSVM, and results obtained by CSVM are less likely to be better than those obtained by BCSVM. Moreover, BCSVM only reduces the error between CSVM and direct training, but does not completely solve the error problem caused by group training. At the same time, the paper does not give the specific value of error reduction. Therefore, the following work focuses on the performance optimization of group training. The model obtained by group training is the same as the model obtained by direct training, so as to solve the problem that the training time of SVM is too long and there will be errors in grouping.

\bibliography{sn-bibliography}


\end{document}